\documentclass[conference,9pt]{IEEEtran}
\IEEEoverridecommandlockouts
\usepackage[table,xcdraw]{xcolor}
\usepackage{cite}
\usepackage{amsmath,amssymb,amsfonts}
\usepackage{bm}
\usepackage{algorithmic}
\usepackage{graphicx}
\usepackage{textcomp}
\usepackage{xcolor}
\usepackage[utf8]{inputenc} 
\usepackage[T1]{fontenc}    
\usepackage{multirow}
\usepackage{booktabs}
\usepackage[hyphens]{url}
\usepackage{enumitem}
\usepackage[hidelinks]{hyperref}
\usepackage{booktabs}
\usepackage{subcaption}
\usepackage{caption}
\usepackage{makecell}
\graphicspath{{./figures/}}
\def\BibTeX{{\rm B\kern-.05em{\sc i\kern-.025em b}\kern-.08em
    T\kern-.1667em\lower.7ex\hbox{E}\kern-.125emX}}
\begin{document}

\newcommand{\todo}[1]{{\color{red} #1}}
\newcommand{\revise}[1]{{\color{blue} #1}}

\title{Ralts: \underline{R}obust \underline{A}ggregation for Enhancing Graph Neura\underline{l} Ne\underline{t}work Re\underline{s}ilience on Bit-flip Errors}

\author{Wencheng Zou, Nan Wu\\
George Washington University\\
wenchengz@gwu.edu, nan.wu@gwu.edu}
\maketitle

\begin{abstract}
Graph neural networks (GNNs) have been widely applied in safety-critical applications, such as financial and medical networks, in which compromised predictions may cause catastrophic consequences.
While existing research on GNN robustness has primarily focused on software-level threats, hardware-induced faults and errors remain largely underexplored.
As hardware systems progress toward advanced technology nodes to meet high performance and energy efficiency demands, they become increasingly susceptible to transient faults, which can cause bit flips and silent data corruption, a prominent issue observed by major technology companies (e.g., Meta and Google).
In response, we first present a comprehensive analysis of GNN robustness against bit-flip errors, aiming to reveal system-level optimization opportunities for future reliable and efficient GNN systems.
Second, we propose Ralts, a generalizable and lightweight solution to bolster GNN resilience to bit-flip errors.
Specifically, Ralts exploits various graph similarity metrics to filter out outliers and recover compromised graph topology, and incorporates these protective techniques directly into aggregation functions to support any message-passing GNNs.
Evaluation results demonstrate that Ralts effectively enhances GNN robustness across a range of GNN models, graph datasets, error patterns, and both dense and sparse architectures.
On average, under a BER of $3\times10^{-5}$, these robust aggregation functions improve prediction accuracy by at least 20\% when errors present in model weights or node embeddings,  and by at least 10\% when errors occur in adjacency matrices.
Ralts is also optimized to deliver execution efficiency comparable to built-in aggregation functions in PyTorch Geometric.
\end{abstract}

\section{Introduction}

Graph neural networks (GNNs) have achieved remarkable success in a variety of applications~\cite{kipf2016semi,yue2020graph,wu2020comprehensive,wu2021ironman}.
Their adoption has expanded into risk-sensitive areas, such as detecting fraud in financial networks~\cite{amazon_fraud,motie2023financial,duan2024cat}, identifying money laundering~\cite{weber2018scalable}, exposing camouflaged fraudsters~\cite{dou2020enhancing}, detecting malware~\cite{hou2019alphacyber}, and predicting system vulnerability~\cite{alrahis2023graph,chu2024graph}.
In 2023, Federal Trade Commission~\cite{FTC} reported over \$10B fraud losses in the US, a 14\% increase over 2022; another source reported that bank transfer or payment fraud resulted in losses amounting to \$1.59B~\cite{fraud-loss}. 
In this context, \textit{GNN robustness is more critical than ever}, as mispredictions in these safety-critical systems could enable large-scale financial or cyber crimes, resulting in substantial losses for individuals and institutions.


As hardware systems advance into nanometer technology nodes, they become more sensitive to process, voltage, temperature, and aging (PVTA) variations, significantly increasing the risk of hardware failures and data corruption.
For instance, transient errors (e.g., timing errors) that escape detection during manufacturing tests, have become more prevalent~\cite{shafique2020robust}.
Common hardware errors include radiation-induced soft errors~\cite{hazucha2003neutron}, variation-induced timing errors~\cite{ernst2004razor,shafique2020robust}, and errors from voltage scaling~\cite{chandra2008impact,zhang2018thundervolt}.
Such errors often cause bit flips during computation, resulting in silent data corruption (SDC), a growing issue in large-scale infrastructures at major technology companies like Meta~\cite{dixit2022detecting} and Google~\cite{bacon2022detection}.
With increasing interest in GNNs and their applications, it is essential to conduct a thorough analysis of GNN robustness against hardware errors.

\begin{figure}
    \centering
    \includegraphics[width=0.96\linewidth]{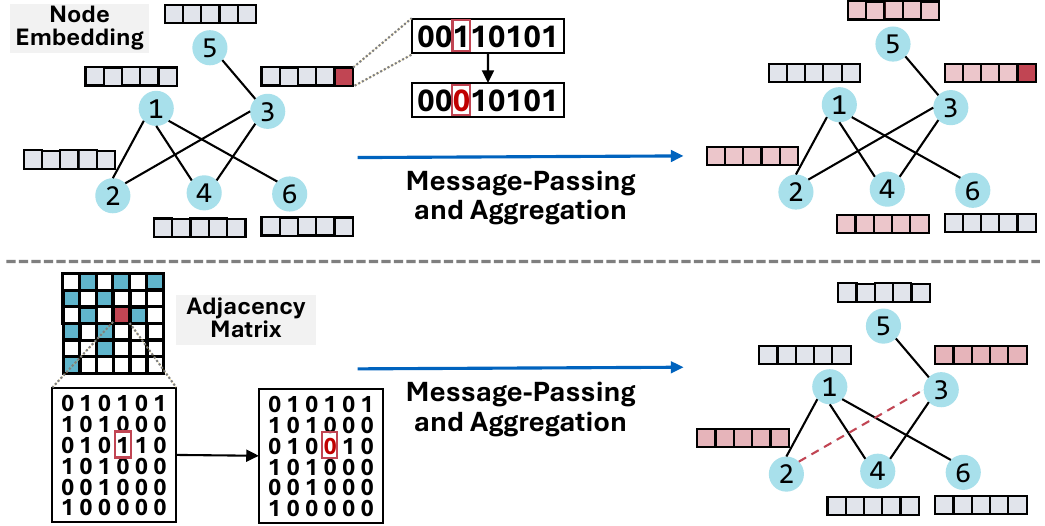}
    \caption{Bit errors in node embeddings (top subfigure) and adjacency matrices (bottom subfigure) propagate to neighboring nodes within one iteration of message-passing. Values with bit errors are marked in red, and affected node embeddings are shaded in light red. 
    Since spatial GNNs rely on topological information through the message-passing mechanism, their robustness behavior differs from that of DNNs with regular inputs.}
    \label{fig:errors_in_gnn}
\end{figure}

The exploration of GNN resilience on hardware errors has left two major gaps unaddressed.
(1) From the perspective of GNN robustness assessment, first, the irregular structure of graph data creates unique robustness behaviors in GNNs, contrasting with the extensive studies on deep neural network (DNN) resilience to bit errors~\cite{jiao2017assessment,whatmough2018dnn,reagen2018ares,stutz2021bit}.
As shown in Fig.~\ref{fig:errors_in_gnn}, bit errors in GNNs can stem from node embeddings, adjacency matrices, and model weights, leading to degraded GNN performance.
Second, existing GNN resilience studies ignore either hardware-induced errors~\cite{jin2021adversarial,sun2022adversarial} or vulnerabilities tied to graph topology~\cite{wang2022pygfi}.
Thus, it is imperative to conduct \textit{a comprehensive analysis of GNN robustness against hardware bit errors}, which could reveal new optimization opportunities for reliable and efficient GNN systems.
(2) From the protection perspective, current methods for safeguarding graphs or GNNs introduce high computational overhead, ignoring scalability and execution efficiency~\cite{sun2022adversarial}.
For instance, many gradient-based graph recovery methods require either complex matrix manipulations (e.g., singular value decomposition to enforce low-rank adjacency matrices~\cite{jin2020graph}) or constrained projected gradient descent (PGD)~\cite{xu2019topology,geisler2021robustness}, effective but computationally expensive.
Thus, it is crucial to develop \textit{general, robust, and hardware-efficient protection mechanisms for GNNs}.

In response, we propose Ralts, robust aggregation designed to enhance GNN resilience against bit-flip errors.
Specifically, Ralts exploits various graph similarity metrics to filter out outliers and recover compromised graph topology.
By integrating these protective techniques directly into aggregation functions, Ralts can be easily applied across diverse GNN architectures and takes advantage of the efficient execution of the message-passing mechanism.
Ralts is optimized to achieve execution efficiency comparable to built-in aggregation functions in PyTorch Geometric (PyG)~\cite{fey2019fast}.
The contributions are summarized as follows.
\begin{itemize}
    \item \textbf{Comprehensive analysis of GNN robustness on bit-flip errors.} We systematically analyze the impact of bit errors on GNN performance, aiming to uncover system-level optimization opportunities for building reliable and efficient GNN systems.
    \item \textbf{Robust, general, and lightweight protection mechanisms for GNNs.} We propose three aggregation functions that exploit various graph similarity metrics to filter out outlier values and recover graph topology, significantly boosting GNN robustness against bit errors. Such robust aggregation functions can be easily integrated into any message-passing GNNs, with high execution efficiency.
    \item \textbf{Evaluation on robustness and efficiency.} 
    Our results demonstrate that Ralts serves as a generalizable approach for enhancing GNN robustness across different GNN models, graph datasets, error patterns, and both dense and sparse architectures. 
    On average, under a bit error rate (BER) of $3\times10^{-5}$, these robust aggregation functions improve prediction accuracy by at least 15\%.
    Profiling results indicate that Ralts achieves execution efficiency on par with PyG built-in functions.
\end{itemize}

\section{Background and Motivations}

\subsection{GNNs: Efficiency and Robustness}
Spatial GNNs leverage graph topology as inductive biases~\cite{kipf2016semi,xu2018powerful} and employ the message-passing mechanism as their backbone~\cite{wu2020comprehensive}.
The computation can be generally expressed as:
\begin{equation}
h_{v}^{(l+1)}=\gamma(h_{v}^{(l)},\underset{u\in\mathcal{N}_v}{\mathcal{A}}(h_{u}^{(l)})).
\end{equation}
Here, $h_{v}^{(l)}$ is the node embedding of node $v$ at layer $l$; $\mathcal{N}_v$ denotes the neighboring nodes of $v$; $\mathcal{A}(\cdot)$ is a permutation-invariant aggregation function; $\gamma(\cdot)$ is a differentiable message transformation function.

As GNNs are increasingly applied in safety-critical scenarios, both their efficiency and robustness are essential.
Existing efficient hardware systems for GNNs~\cite{abadal2021computing} primarily focus on system performance, while most GNN robustness analyses are conducted from a software adversarial perspective~\cite{jin2021adversarial, sun2022adversarial}, leaving GNN robustness against hardware-induced errors unexplored.
While some adversarial defense techniques can be adapted to mitigate hardware errors, they often introduce substantial computational overhead, sacrificing efficiency for robustness.
For example, Pro-GNN~\cite{jin2020graph} reconstructs graph structures through feature smoothing and by enforcing a low-rank and sparse adjacency matrix using singular value decomposition;
test-time graph transformation~\cite{jinempowering} refines graph topology by PGD~\cite{xu2019topology,geisler2021robustness}.
These protection mechanisms face challenges in scalability and execution efficiency~\cite{sun2022adversarial}.

\subsection{Bit Errors in Hardware}
\begin{figure}
    \centering
    \includegraphics[width=0.98\linewidth]{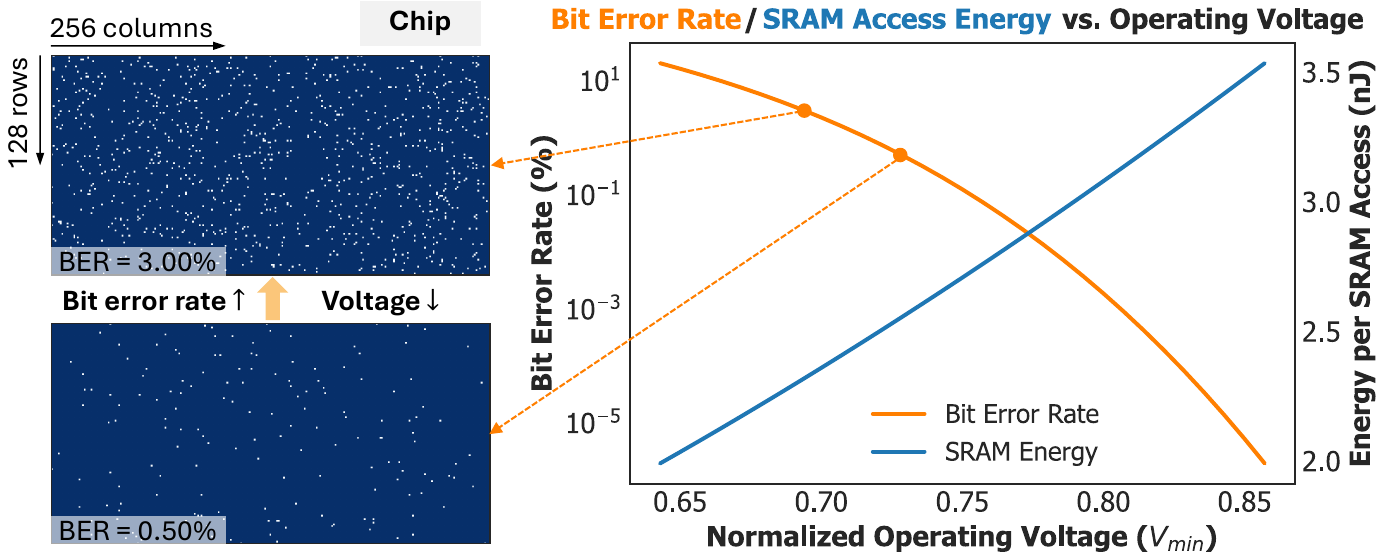}
    \caption{Bit errors and energy consumption in memory with voltage scaling. SRAM bit error rate (\textcolor{orange}{orange}, left y-axis) increases while energy consumption (\textcolor{blue}{blue}, right y-axis) decreases as the supply voltage reduces in 14nm FinFET SRAM arrays~\cite{chandramoorthy2019resilient}, redrawn from data in~\cite{wan2023berry,wan2024mulberry}. The voltage (x-axis) is normalized to $V_{min}$, the lowest measured voltage at which no bit errors are observed. 
    The two subfigures on the left show a random spatial error pattern in a cross-section of the memory array, reproduced from~\cite{chandramoorthy2019resilient,stutz2021bit}.}
    \label{fig:ber}
\end{figure}

With technology scaling to the nanoscale, silicon-based hardware has become more susceptible to transient errors due to PVTA variations, such as soft errors in memory or on-chip buffers~\cite{chandra2008impact, zhang2018thundervolt} and timing errors in processing elements~\cite{ernst2004razor, shafique2020robust}.
As the demand for sustainability and power efficiency grows, applications often operate at lower voltages, exacerbating bit cell variations.
Fig.~\ref{fig:ber} illustrates an exponential increase in bit error rates (BERs) with voltage scaling in a 14nm FinFET SRAM chip~\cite{chandramoorthy2019resilient}, along with a sample spatial distribution of bit errors within a segment of tested memory arrays. 
These error locations are random and independent across different chips and arrays~\cite{ganapathy2017characterizing, kim2018matic, stutz2021bit}, impacting accelerator memories that store and update weights and intermediate results.
Though error correction codes (ECC)~\cite{macwilliams1977theory} offer some protection, they have two primary limitations: first, standard ECC may not fully mitigate errors when multiple faulty bits occur within a single memory word; second, the hardware overhead of ECC rises significantly with the number of bits to be corrected: for example, protecting a 64-bit data word incurs overheads of 11\%, 22\%, and 32\% for single-bit, double-bit, and triple-bit corrections, respectively~\cite{2008161}.

\subsection{Motivations}
Given the importance of GNN robustness and the lack of investigation into hardware-induced errors, we aim to provide a comprehensive analysis of GNN robustness in the presence of bit errors. 
By characterizing GNN tolerance to hardware errors, we can identify optimization opportunities across various levels of system abstraction to better balance efficiency and reliability requirements under different scenarios, for instance, applying voltage scaling at the hardware architecture level, adjusting ECC costs at the device level, and exploring emerging technologies and aggressive design rules at the fabrication level.

Considering the high overheads of adversarial defense techniques and ECC, there is a strong need for \textit{a lightweight and general protection mechanism for GNNs} that maintains both efficiency and robustness. 
The proposed method, Ralts, integrates protection mechanisms directly into aggregation functions, making it readily adaptable to message-passing GNN acceleration systems~\cite{abadal2021computing}. 
As an algorithm-level robustness technique, Ralts is complementary to circuit-level timing error correction~\cite{zhang2018thundervolt, pandey2019greentpu} and architecture-level dataflow optimization~\cite{zhang2023read}, and can be combined with these techniques.
\section{Bit Error Injection in GNNs}

\begin{figure}[b]
    \centering
    \includegraphics[width=0.95\linewidth]{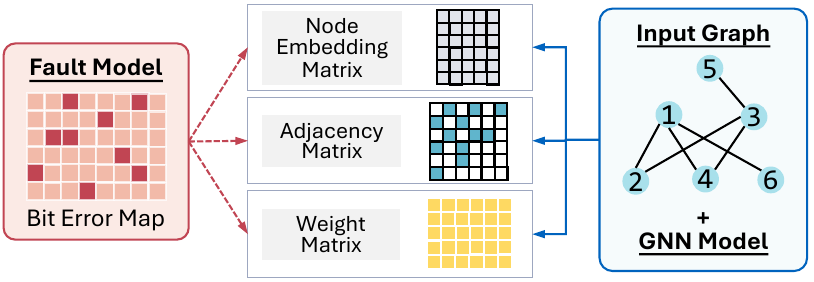}
    \caption{Bit error injection for analyzing GNN robustness, where bit flips can be introduced into node embeddings, adjacency matrices, and model weights.}
    \label{fig:fault_injection}
\end{figure}
\begin{figure*}[t]
    \centering
    \includegraphics[width=\linewidth]{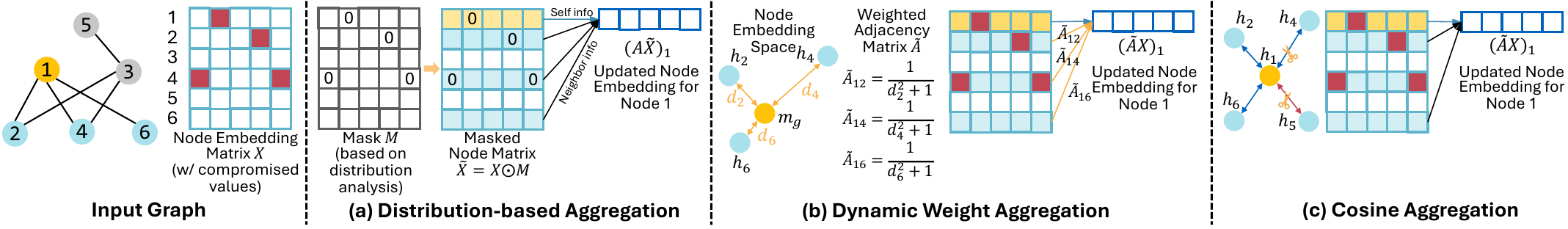}
    \caption{Robust aggregation in GNNs: (a) distribution-based aggregation trims extreme values using distribution analysis; (b) dynamic weight aggregation assigns weights based on the distance to a learnable center; (c) cosine aggregation simultaneously reconstructs graph structure and trims extreme values, where the red edge is a compromised edge by adversaries. Note that node 1 is taken as the target node being updated as an example.}
    \label{fig:aggregation}
\end{figure*}

Following the approach of existing DNN robustness studies~\cite{jiao2017assessment,reagen2018ares,stutz2021bit,kim2018matic}, we focus on memory errors and adopt a random fault model based on silicon characterizations~\cite{chandramoorthy2019resilient,ganapathy2017characterizing,kim2018matic}. 
In this model, each memory bit has a probability $p$ to be flipped, i.e., BER, which aligns with practical scenarios where memory faults are not detected or corrected by ECC; each bit flips independently of the others. 
Such bit flips directly impact the stored values.

To thoroughly analyze GNN robustness against bit errors, we inject bit flips in node embeddings, model weights, and adjacency matrices of GNNs, as shown in Fig.~\ref{fig:fault_injection}.
In addition to dense GNN models, we also explore the robustness of sparse GNNs, as model sparsification is considered as an effective method to enhance DNN robustness against bit flips~\cite{guo2018sparse,gao2020reliability}.
While our primary focus is on random memory errors, the developed bit error injection framework can be accommodated for other hardware fault models.
\section{Robust Aggregation in GNNs}

The aggregation function in message-passing GNNs collects information from neighboring nodes using operations such as mean, max, or sum. However, these standard operations are often sensitive to data perturbations~\cite{liang2021understanding}. To enhance robustness against bit-flip errors that may compromise both graph data and GNN model weights, we propose three aggregation techniques that incorporate graph similarity metrics to filter out extreme values and recover structural integrity, as illustrated in Fig.~\ref{fig:aggregation}.
These robust aggregation functions are inherently aligned with the properties of graph-structured data: nodes with high similarity are more likely to be connected~\cite{zhang2018link,li2024evaluating}. For instance, in clean graphs, neighboring nodes typically exhibit similar feature distributions~\cite{mcpherson2001birds}; likewise, in citation networks, connected papers often cover related research topics~\cite{kipf2016semi}.

\subsection{Aggregation based on Distribution Analysis}

Intuitively, removing outliers or extreme values introduced by bit-flip errors helps prevent the propagation of corrupted information to neighboring nodes, thereby enhancing robustness against data perturbations. Empirical observations show that node embeddings in GNNs often follow certain distribution patterns, which enables the identification and filtering of compromised features or activations that deviate significantly from expected ranges.

\begin{figure}
    \centering
    \includegraphics[width=\linewidth]{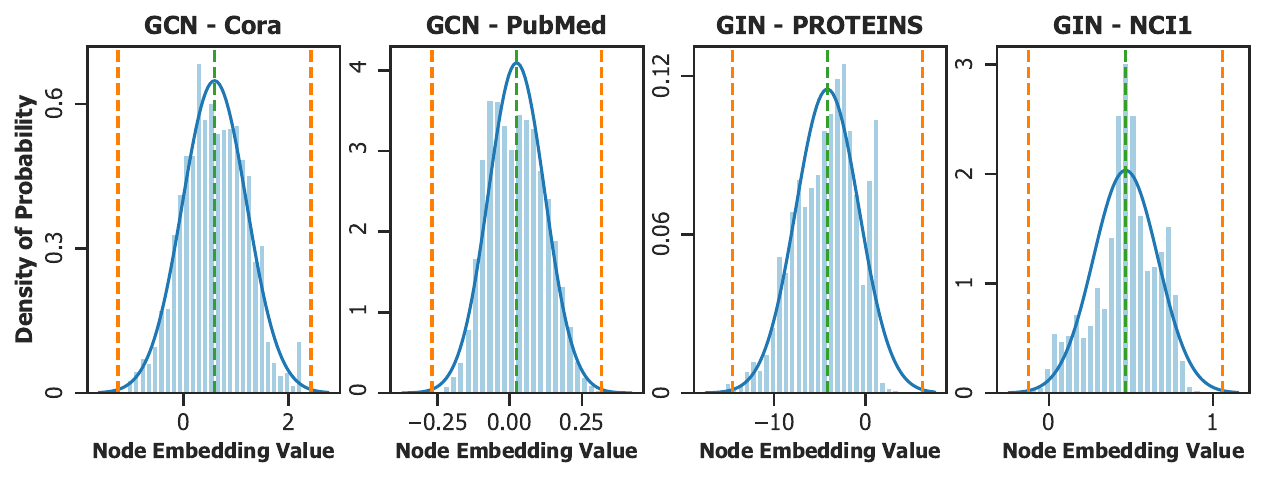}
    \caption{Distribution of the first element in the first-layer node embeddings of GCN and GIN on Cora, PubMed, PROTEINS, and NCI1 datasets. Notably, all of them exhibit distributions resembling Gaussian distributions (fitted in blue lines), with mean $\mu$ and $\mu\pm3\sigma$ marked in dashed lines.}
    \label{fig:distribution}
\end{figure}

According to the Central Limit Theorem, under appropriate conditions, the distribution of sample means approximates a normal (i.e., Gaussian) distribution, when the sample size is sufficiently large.
Since GNN aggregation functions typically perform neighborhood averaging, the values in each dimension of node embeddings are expected to exhibit near-Gaussian behavior. In Gaussian distributions, approximately 97.33\% of data values lie within three standard deviations from the mean. Thus, values falling outside this range can be regarded as anomalies.
Fig.~\ref{fig:distribution} illustrates the distributions of node embeddings across various datasets and GNN models.
While not all of them strictly follow Gaussian distributions, we still can apply a threshold-based mechanism to trim values deviating multiple standard deviations from the mean.

The proposed distribution-based aggregation operates as follows. During training, the distribution of each element in node embeddings is analyzed to estimate the mean $\mu$ and standard deviation $\sigma$, without trimming any values. 
During inference, the aggregation function sets a confidence interval of $(\mu-a\sigma, \mu+b\sigma)$, where $a$ and $b$ are positive hyperparameters adjusted for different datasets or GNNs; only values within this interval are averaged, while those outside the range are discarded, as illustrated in Fig.~\ref{fig:aggregation}(a).

\subsection{Aggregation based on Dynamic Weight}
In scenarios where extreme values may still contain valuable information, we propose an alternative weighted mean aggregation strategy that attenuates the influence of potential outliers rather than discarding them entirely. This method assigns lower weights to node embeddings that are likely to be compromised, based on their distance from a learned central representation of the graph.
Specifically, we introduce a learnable center embedding $m_g$ for the graph, which serves as a reference point. For each node $u$, we compute the squared Euclidean distance from its node embedding $h_u$ and the center embedding as $d_u^2 = ||h_u - m_g||^2$. Intuitively, embeddings that are farther from the center are less similar to the typical nodes in the graph and are thus given lower weights during aggregation.
The weight assigned to node $u$ is then defined as:
\begin{equation}
    \text{Weight}_u=\frac{1}{d^2_{u}+1}.
    \label{eq:dynamic_weight}
\end{equation}

Fig.~\ref{fig:aggregation}(b) shows an example computation flow of the dynamic weight aggregation. During training, the graph center embedding $m_g$ is learned jointly with the model parameters via backpropagation, capturing the central tendency of node embeddings across the dataset. During inference, the aggregation function dynamically assigns weights to neighboring nodes according to their distances from the center embedding $m_g$, as in Equation (\ref{eq:dynamic_weight}), and then performs a weighted mean over the neighbors.
This formulation ensures that all nodes contribute to the aggregation, but nodes with larger deviations from the center have proportionally reduced influence, providing a smooth trade-off between robustness and information retention.

\subsection{Aggregation based on Cosine Similarity}

Distribution-based aggregation and dynamic weight aggregation primarily address outliers in node embeddings and model weights. However, since bit-flip errors can also affect the adjacency matrix, as illustrated in Fig.~\ref{fig:errors_in_gnn}, we propose an additional aggregation function aimed at graph topology recovery, leveraging cosine similarity between node pairs.
The key idea is that structurally similar nodes are more likely to be connected, allowing for the reconstruction of corrupted graph structures by pruning spurious edges with low similarity.
Specifically, for any two vectors $\bm{x}_i$ and $\bm{x}_j$, cosine similarity is defined as: 
\begin{equation} 
\text{cosine similarity}(\bm{x}_i, \bm{x}_j) = {\frac {\bm{x}_i\cdot \bm{x}_j} {||\bm{x}_i||~||\bm{x}_j||}}. \label{eq:cosine} 
\end{equation}
For each node $v$, we compute cosine similarity with all its neighbors $u \in \mathcal{N}_v$. A tunable threshold parameter $\alpha$ is introduced to accommodate variations across graph datasets and GNN models. If the similarity score between $v$ and a neighbor $u$ falls below $\alpha$, the edge $(v, u)$ is considered spurious and removed.
In addition to restoring corrupted graph topology, this approach can partially mitigate the impact of corrupted node features, further enhancing robustness to bit-level errors.

\subsection{Combination of Aggregation Functions}
Inspired by the insight that combining multiple aggregation functions can better capture local graph structures~\cite{corso2020principal}, we propose integrating multiple robust aggregation techniques, each scaled by a learnable scalar, to further improve the robustness of GNNs.
Each aggregation method targets different aspects of compromised data: distribution-based aggregation filters out statistical outliers, dynamic weight aggregation down-weights distant embeddings, and cosine similarity-based aggregation recovers corrupted graph topology.
By leveraging their complementary strengths, this ensemble approach enhances resilience against a wide range of bit-flip errors.
Moreover, since these aggregation functions are independent, they can be computed in parallel, incurring minimal additional latency during inference.
\section{Experiment Setup}

To comprehensively analyze GNN robustness against bit-flip errors and demonstrate the generalizability of the proposed robust aggregation functions, we conduct extensive experiments to evaluate their effectiveness across various GNN models, diverse graph datasets, different error patterns, and both dense and sparse model configurations. The detailed experimental settings are summarized as follows.

\begin{itemize}
    \item \textbf{GNN models.} We conduct the experiment with three widely used GNN models: graph convolutional network (GCN)~\cite{kipf2016semi}, graph attention network (GAT)~\cite{velivckovic2017graph}, and graph isomorphism network (GIN)~\cite{xu2018powerful}.
    \item \textbf{Datasets.} We first evaluate the proposed robust aggregation functions on six graph datasets including both node-level and graph-level tasks: Cora~\cite{yang2016revisiting}, PubMed~\cite{pubmed}, Citeseer~\cite{yang2016revisiting}, MUTAG~\cite{kriege2012subgraph}, PROTEINS~\cite{proteins}, and NCI1~\cite{nr}. To assess scalability and effectiveness on larger and denser graphs, we further include two datasets from the Open Graph Benchmark (OGB): ogbn-arxiv~\cite{wang2020microsoft} and ogbn-products~\cite{Bhatia16}.
    Detailed statistics of all datasets are presented in Table~\ref{tab:dataset}.
    \item \textbf{Error injection.} Bit-flip errors are injected into node embeddings, adjacency matrices, and model weights, under varying BERs to simulate different levels of perturbation.
    \item \textbf{Baseline aggregation functions.} The proposed robust aggregation functions are compared against several baselines:
    \begin{itemize}
        \item PyG built-in aggregation methods, such as max and mean;
        \item existing aggregation techniques designed for improved robustness to data perturbations, including median aggregation and trimmed mean aggregation~\cite{liang2021understanding}, as well as soft median aggregation~\cite{geisler2021robustness};
        \item other outlier-trimming mechanisms in GNNs aimed at mitigating bit-flip errors, such as topology-aware activation clipping~\cite{wang2022pygfi}.
    \end{itemize}
    \item \textbf{Implementation details.} All GNN models are implemented using PyG~\cite{fey2019fast}, with the fault model and error injection mechanisms developed in Python. Hyperparameters for each GNN model are carefully tuned to achieve prediction accuracy comparable to state-of-the-art benchmarks. To ensure statistical reliability, the reported results are averaged over five different random seeds, with 10 independent runs per seed.
    \item \textbf{Model sparsification.} We further evaluate the robustness of sparse GNNs against bit-flip errors. We employ surrogate Lagrangian relaxation-based weight pruning~\cite{peng2022towards} to obtain models with varying sparsity levels: 15\%, 30\%, and 60\%.
    \item \textbf{Profiling platform.} To evaluate the efficiency of different aggregation functions, we profile their execution latency on a server with 28-core Intel Xeon Ice Lake Gold 6330 CPUs and NVIDIA RTX A6000 GPUs.
\end{itemize}

\begin{table}[t]
    \centering
    \caption{Dataset statistics.}
    \setlength{\tabcolsep}{4.6pt}
    \renewcommand{\arraystretch}{1.1}
    \begin{tabular}{c|cccc}\toprule
         \textbf{Dataset} & \textbf{\# Graphs} & \textbf{\# Nodes} & \textbf{\# Edges} & \textbf{Task} \\ \midrule
        Cora~\cite{yang2016revisiting} & 1 & 2,708 & 10,556 & Node-level \\
        PubMed~\cite{pubmed} & 1 & 19,717 & 88,648  & Node-level \\
        Citeseer~\cite{yang2016revisiting} & 1& 3,327 & 9,104  & Node-level \\
        MUTAG~\cite{kriege2012subgraph} & 188 & 27,163 & 148,454 & Graph-level  \\
        PROTEINS~\cite{proteins} & 1113 & 43,471 & 81,049 & Graph-level  \\
        NCI1~\cite{nr} &  4110 & 122,747 &  132,753 & Graph-level \\
        ogbn-arxiv~\cite{wang2020microsoft} & 1 & 169,343 & 1,166,243 & Node-level \\
        ogbn-products ~\cite{Bhatia16} & 1 & 2,449,029 & 61,859,140 & Node-level \\ \bottomrule
    \end{tabular}
    \label{tab:dataset}
\end{table}



\section{Analysis of GNN Robustness on Bit-flip Errors}

\begin{figure}[t]
    \centering
    \includegraphics[width=\linewidth]{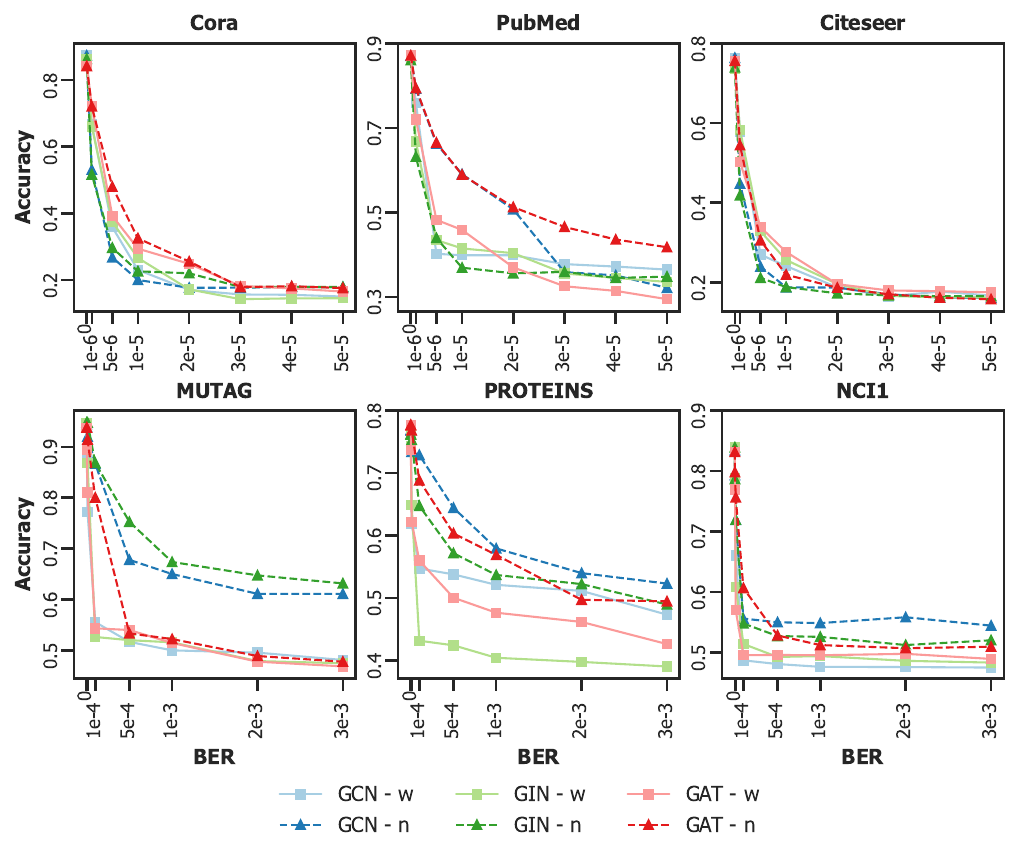}
    \caption{GNN prediction accuracy under different intensities of bit-flip errors in model weights (denoted by `-w' with square markers) and node embeddings (denoted by `-n' with triangle markers).}
    \label{fig:gnn_flip}
\end{figure}

We evaluate the robustness of GNNs to bit-flip errors by injecting faults into model weights, node embeddings, and adjacency matrices.
Fig.~\ref{fig:gnn_flip} presents the prediction accuracy of various GNN models under bit-flip errors in model weights and node embeddings across multiple datasets.
Three key observations can be made.
\underline{\textit{First}}, GNNs generally maintain stable performance for BERs below $10^{-7} \sim 10^{-6}$, when only a single data type (e.g., weights or embeddings) is affected. This indicates that, for applications where ultra-high reliability is not critical, aggressive system-level optimizations, such as voltage scaling, can be leveraged to improve energy efficiency.
\underline{\textit{Second}}, GNNs exhibit higher sensitivity to bit-flip errors in model weights compared to those in node embeddings. This is likely due to the direct influence of weights on all model computations, while corrupted embeddings can be partially mitigated by neighborhood aggregation.
\underline{\textit{Third}}, smaller graphs (e.g., MUTAG, PROTEINS, NCI1) demonstrate greater resilience to bit-flip errors, which can be attributed to the reduced data volume in both model weights and embeddings.

\begin{figure}[t]
    \centering
    \includegraphics[width=\linewidth]{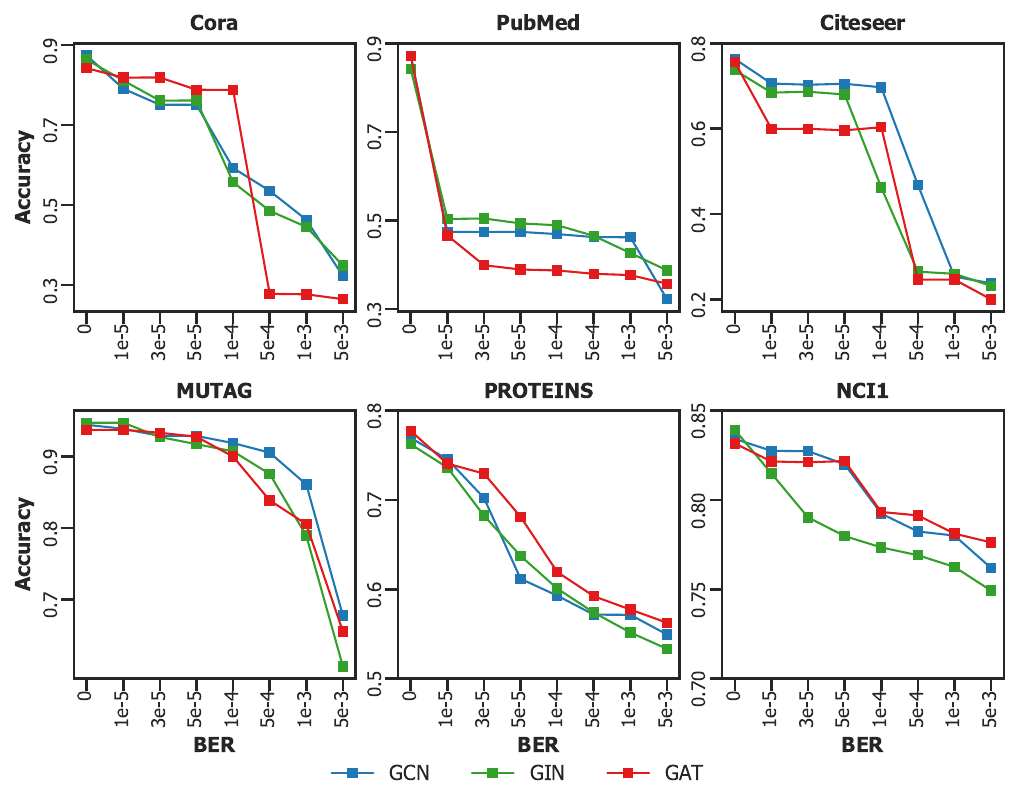}
    \caption{GNN prediction accuracy under different intensities of bit-flip errors in adjacency matrices. }
    \label{fig:gnn_flip_edge}
\end{figure}

For errors in adjacency matrices, the impact on model accuracy varies across datasets, as shown in Fig.~\ref{fig:gnn_flip_edge}. This aligns with the intuition that adjacency matrices encode graph structural information, and different graphs exhibit varying levels of robustness to topology perturbations.

\section{Evaluation on Robust Aggregation}

To comprehensively evaluate the effectiveness of Ralts, we investigate the following four aspects.
\begin{itemize}
    \item \textbf{Effectiveness of outlier trimming with high information retention.} We examine the proportion of data in the final GNN layer impacted by bit-flip errors, along with the percentage of data discarded by the robust aggregation functions, to evaluate their effectiveness in filtering out corrupted values while retaining informative features.
    \item \textbf{Accuracy improvement under bit-flip errors.} We quantify the improvement in prediction accuracy provided by the proposed robust aggregation techniques, when GNNs are exposed to various levels of bit-flip errors.
    \item \textbf{Consistency across GNN structures.} We assess whether the robustness gains provided by these aggregation functions are consistent across both dense and sparse GNN models, ensuring generalizability.
    \item \textbf{Trade-off between robustness and efficiency.} We evaluate the runtime overhead introduced by robust aggregation, through profiling execution latency and analyzing computational cost, aiming to assess whether the robustness improvements incur significant performance degradation.
\end{itemize}

\subsection{Preserving Information While Eliminating Outliers}

\begin{figure}[t]
    \centering
    \begin{subfigure}[]{0.5\textwidth}
        \centering
        \includegraphics[width=\textwidth]{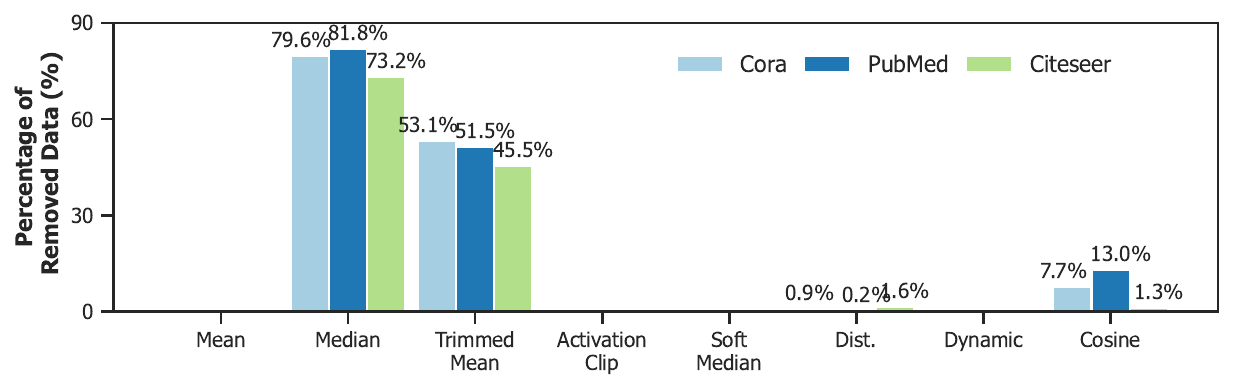}
        \caption{The percentage of node embeddings discarded by different aggregation functions in the absence of bit-flip errors, using GCN as an example. Mean, soft median, and dynamic weight aggregations, as well as activation clip, do not trim any node embeddings.
        Typically, a higher removal rate indicates more information loss.}
        \label{fig:removed_data_a}
    \end{subfigure}
    \hfill
    \begin{subfigure}[]{0.5\textwidth}
        \centering
        \includegraphics[width=\textwidth]{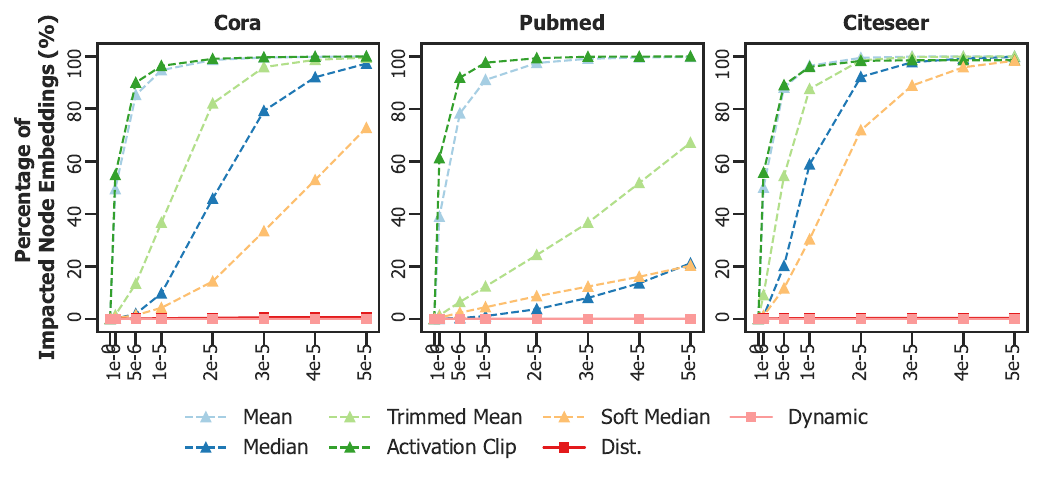}
        \caption{The percentage of node embeddings affected in the final GCN layer under varying BERs when using different aggregation functions. A lower percentage indicates better robustness.}
        \label{fig:removed_data_b}
    \end{subfigure}
    \caption{Percentage of trimmed node embeddings and affected node embeddings in the last GNN layer across different aggregation functions.}
    \label{fig:removed_data}
\end{figure}

We evaluate the effectiveness of robust aggregation functions by analyzing their ability to eliminate outlier data while retaining meaningful information. 
Using GCN as an example, we compare the proportion of node embeddings discarded by different aggregation strategies in the absence of bit-flip errors, as shown in Fig.~\ref{fig:removed_data_a}.
Notably, aggregation methods such as median and trimmed mean adopt aggressive filtering policies, discarding a substantial portion of data to reduce the impact of potential perturbations.
While this strategy can enhance robustness by suppressing corrupted data, it may also remove valuable information, diminishing the model’s capacity to capture subtle variations in local graph topology and reducing its overall expressiveness.

Fig.~\ref{fig:removed_data_b} presents the percentage of node embeddings affected by bit-flip errors in the final layer of GCN, where a lower percentage indicates stronger robustness.
While existing aggregation methods, such as median, soft median, and trimmed mean, substantially reduce the number of impacted embeddings, they are outperformed by the proposed distribution-based and dynamic weight aggregation techniques. 
Our robust aggregation methods not only trim a negligible or even no amount of data, but also effectively identify compromised embeddings, attenuate the influence of outlier values, and preserve as much meaningful information as possible.

\begin{figure}[t]
    \centering
    \includegraphics[width=\linewidth]{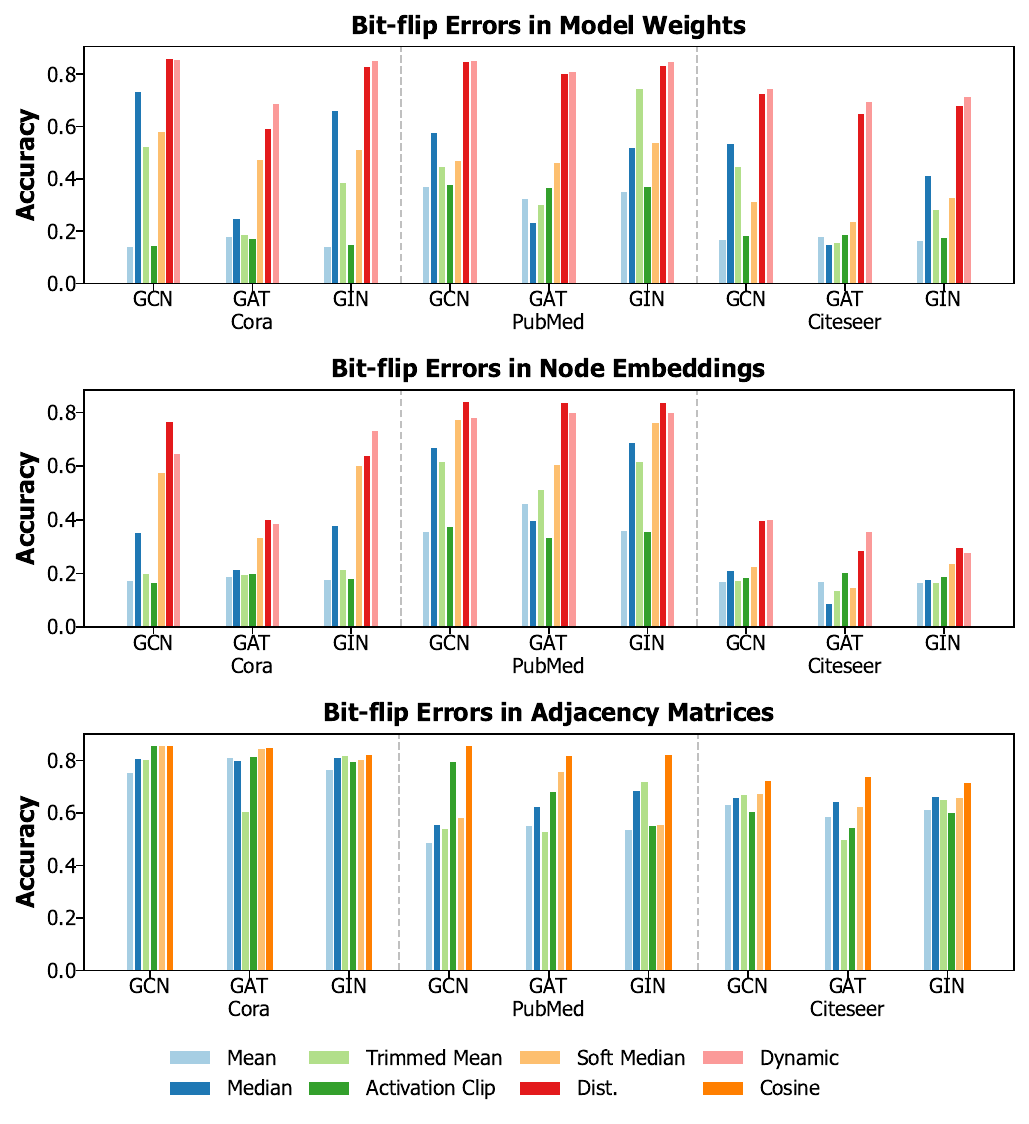}
    \caption{Comparison of GNN prediction accuracy using robust aggregation functions versus existing aggregation methods (mean, median, trimmed mean, activation clip, and soft median) on the Cora, PubMed, and Citeseer datasets. The BER is set as $3\times10^{-5}$. Note that we apply distribution-based aggregation or dynamic weight aggregation for bit flips in model weights/node embeddings, and cosine aggregation for bit flips in adjacency matrices.}
    \label{fig:fixed-ber}
\end{figure}

\begin{figure}[t]
    \centering
    \includegraphics[width=\linewidth]{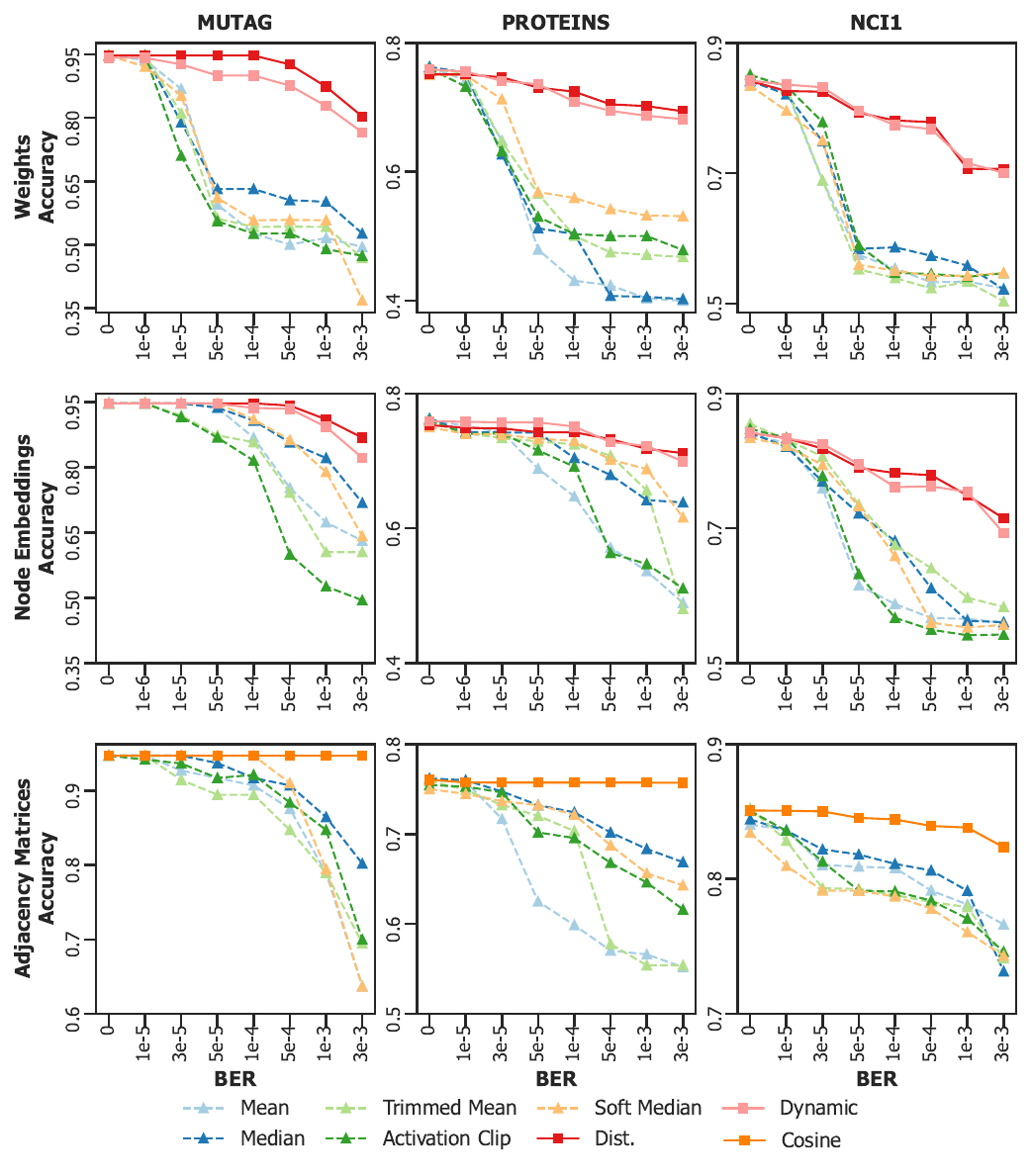}
    \caption{Comparison of GIN prediction accuracy using robust aggregation functions versus baseline aggregation methods (mean, median, trimmed mean, activation clip, and soft median) on graph-level tasks (i.e., MUTAG, PROTEINS, and NCI1). Each column corresponds to a dataset, and each row represents bit-flip errors in model weights, node embeddings, or adjacency matrices.}
    \label{fig:gin-3ds}
\end{figure}

\begin{figure}[h]
    \centering
    \includegraphics[width=\linewidth]{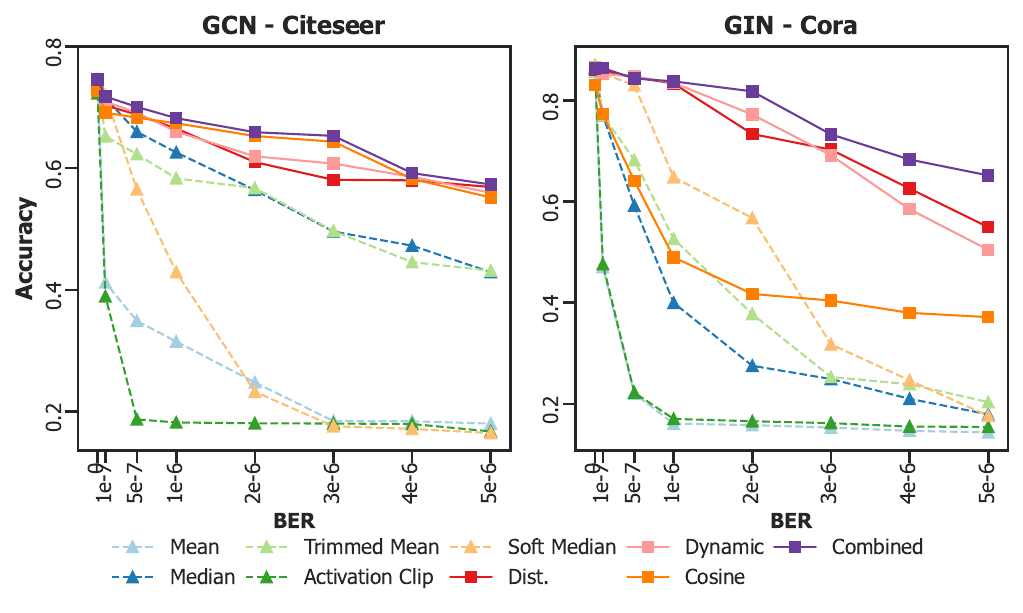}
    \caption{Comparison of GNN prediction accuracy using different aggregation functions, when bit-flip errors happen in model weights, node embeddings, and adjacency matrices. Note that the `Combined' refers to the integration of distribution-based, dynamic weight, and cosine aggregation functions, which can complement each other to further enhance GNN robustness against bit-flip errors. }
    \label{fig:all-flip}
\end{figure}

\subsection{Robustness against Different Bit Errors}
First, we assess the improvement in prediction accuracy achieved by using robust aggregation functions.
Specifically, we inject errors separately into model weights, node embeddings, and adjacency matrices. 
This allows us to assess the effectiveness of distribution-based and dynamic weight aggregation functions in mitigating the impact of outliers in node embeddings or weights, as well as the role of cosine aggregation in recovering graph topologies.
(a) Regarding node-level tasks, as shown in Fig.~\ref{fig:fixed-ber}, using robust aggregation functions can conspicuously improve prediction accuracy under bit-flip errors.
On average, the distribution-based and dynamic weight aggregation functions improve prediction accuracy by 44.1\%, 27.4\%, 32.6\%, 22.2\%, and 43.7\%, compared to mean, median, trimmed mean, soft median aggregation functions, and activation clipping, respectively.
Similarly, the cosine aggregation function achieves improvements of approximately 16.3\%, 10.6\%, 15.2\%, 9.4\%, and 10.6\%, over the same respective baselines.
(b) Regarding graph-level tasks, Fig.~\ref{fig:gin-3ds} shows the prediction accuracy of GIN on three datasets: MUTAG, PROTEINS, and NCI1.
Compared to existing aggregation functions, the proposed distribution-based and dynamic weight aggregation methods demonstrate enhanced robustness against bit-flip errors.
However, their effectiveness appears slightly diminished in graph-level tasks relative to node-level tasks.
One possible explanation is that graph-level tasks depend on graph pooling mechanisms to produce final graph embeddings, which may reduce the influence of individual aggregation functions on overall model performance.
In contrast, cosine aggregation consistently performs well in reconstructing graph topologies.

Second, we conduct experiments in a more challenging scenario, where bit-flip errors are simultaneously introduced in model weights, node embeddings, and adjacency matrices.
As depicted in Fig~\ref{fig:all-flip}, two key takeaways emerge:
(a) GNNs are less stable when perturbations simultaneously occur in model weights, node embeddings, and graph topologies;
(b) although each proposed aggregation function individually enhances robustness, the integration of all of them further enhances robustness, as they rely on distinct graph similarity metrics and complement one another in mitigating errors.

\subsection{Robustness in Sparse GNNs}

\begin{figure}[t]
    \centering
    \includegraphics[width=\linewidth]{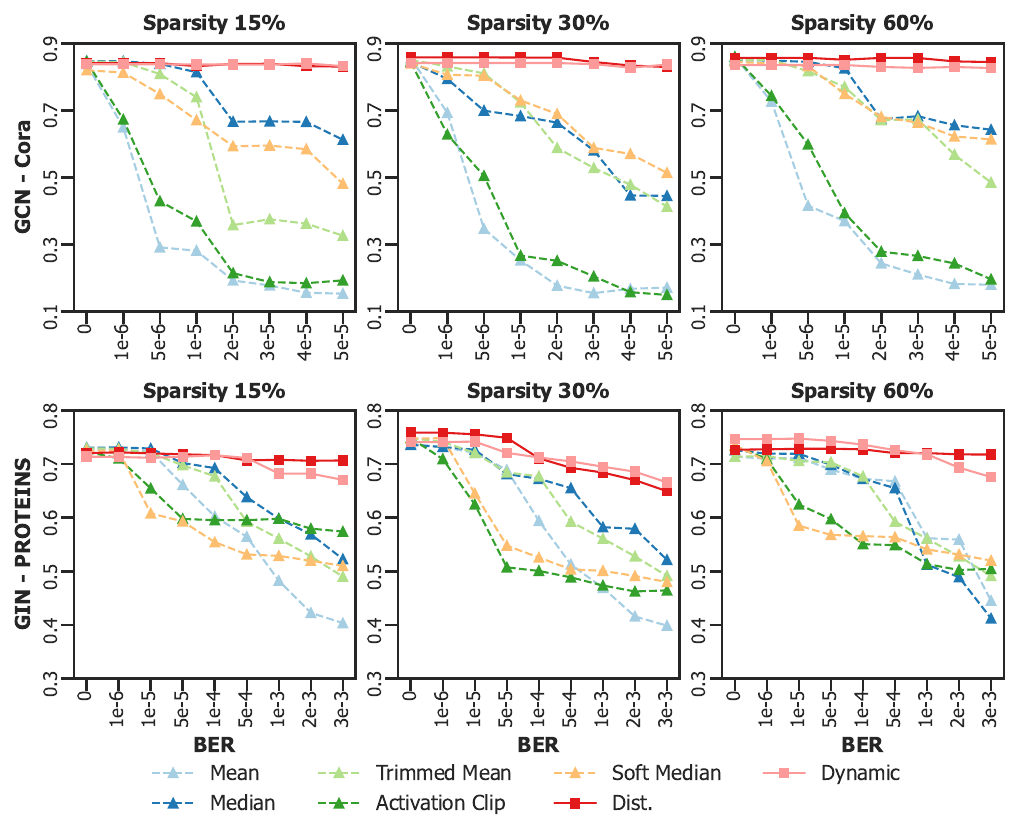}
    \caption{Robust aggregation consistently enhances resilience to bit errors in sparse GNN models. GCN and GIN are pruned to achieve varying levels of sparsity, with bit-flip errors injected into model weights. }
    \label{fig:sparse}
\end{figure}

Although model pruning or sparsification has been shown to enhance DNN robustness~\cite{sehwag2020hydra, chensparsity}, it does not demonstrate the same effects in GNNs, potentially due to the inherent irregularity in graph structures.
As depicted in Fig.~\ref{fig:sparse}, we prune GCN and GIN at different levels of sparsity, and the results indicate that sparse models (using the standard mean aggregation) do not necessarily lead to better robustness against bit-flip errors.
These findings suggest that techniques effective for enhancing DNN robustness may not be directly applicable to GNNs, further highlighting the need for GNN-specific robustness strategies.

In contrast, the proposed robust aggregation functions, grounded in graph similarity metrics, consistently improve GNN resilience to bit-flip errors.
By incorporating statistical insights and leveraging structural properties of graphs, these functions provide stable performance across both dense and sparse GNN models.
Their consistent effectiveness across different sparsity levels underscores their potential as a key strategy for enhancing GNN robustness in the presence of bit-flip errors.

\subsection{Robustness, Execution Efficiency, and Scalability}

\begin{figure}
    \centering
    \includegraphics[width=\linewidth]{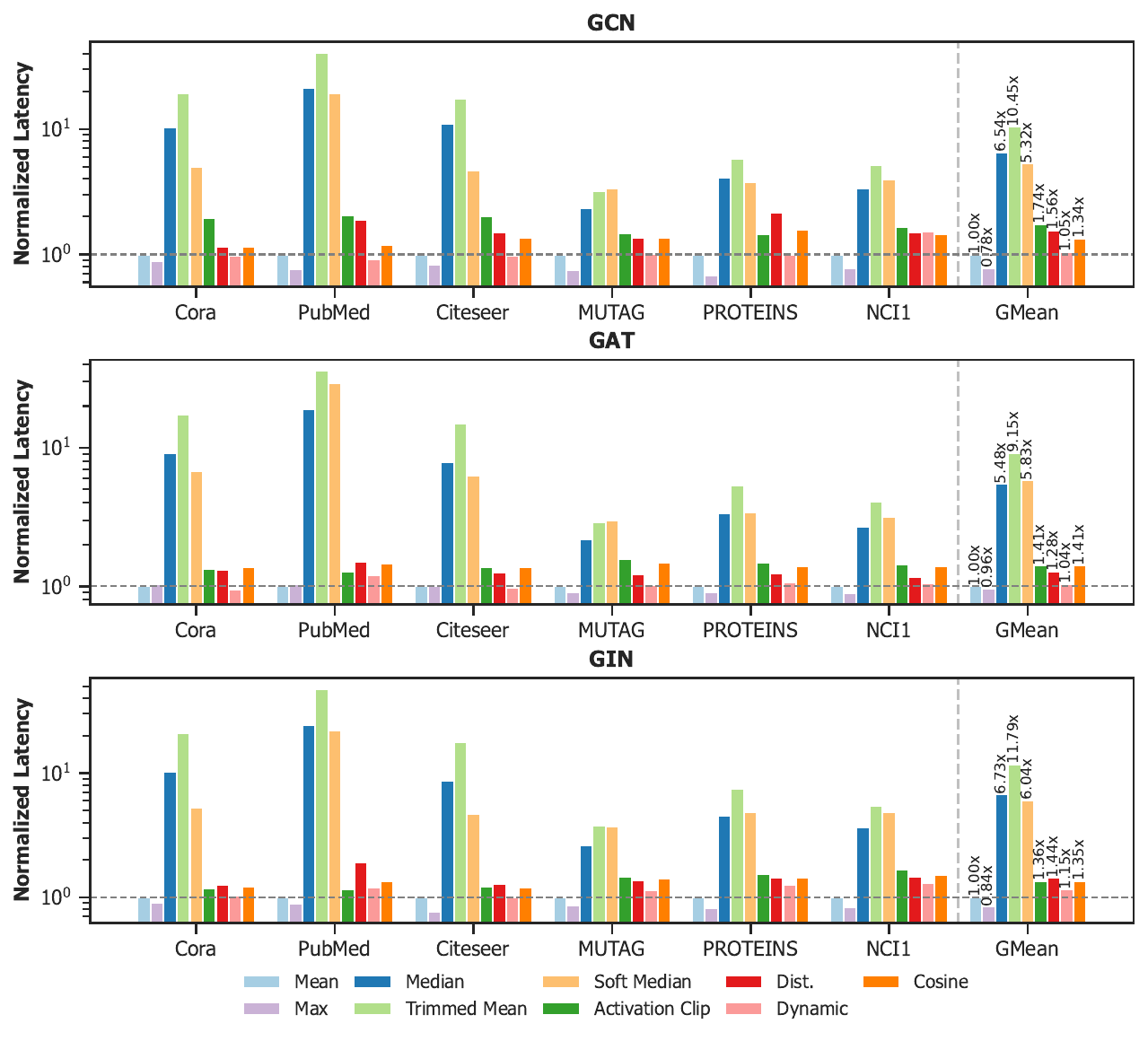}
    \caption{Profiled latency of different aggregation functions across GNN models and graph datasets, normalized to the built-in mean aggregation from PyG. Max aggregation has a similar latency to mean aggregation. Our proposed robust aggregation functions (distribution, dynamic weight, and cosine aggregation functions) achieve a good balance between robustness and execution efficiency. }
    \label{fig:profile}
\end{figure}

To fully unlock the potential of Ralts, it is essential to balance both robustness and execution efficiency.
Fig.~\ref{fig:profile} shows the profiled latency of various aggregation functions across different GNN models and graph datasets.
On geometric average, the distribution-based, dynamic weight, and cosine aggregation functions incur 1.42$\times$, 1.08$\times$, and 1.37$\times$ the latency of the mean aggregation, respectively.
To enhance GNN robustness, introducing additional computations or mechanisms is often inevitable, which may lead to increased execution latency. 
However, our robust aggregation functions are carefully optimized to maintain execution efficiency comparable to PyG built-in aggregation methods.
In contrast, the median, trimmed mean, and soft median aggregation methods are significantly more computationally expensive, 6.22$\times$, 10.41$\times$, and 5.72$\times$ slower than mean aggregation, respectively, while offering only modest gains in prediction accuracy;
although the activation clipping technique achieves a similar level of efficiency as our methods, it provides limited improvements in robustness.

The integration of three robust aggregation functions may introduce slightly higher latency, but it can yield improved robustness under certain scenarios (as demonstrated in Fig.~\ref{fig:all-flip}).
Given that GNN prediction accuracy, execution latency, and robustness are all influenced by the characteristics of graph datasets (e.g., differences in graph topology), these aggregation functions, whether used individually or in combination, offer flexible options to balance the trade-off between efficiency and robustness

\begin{figure}[t]
    \centering
    \includegraphics[width=\linewidth]{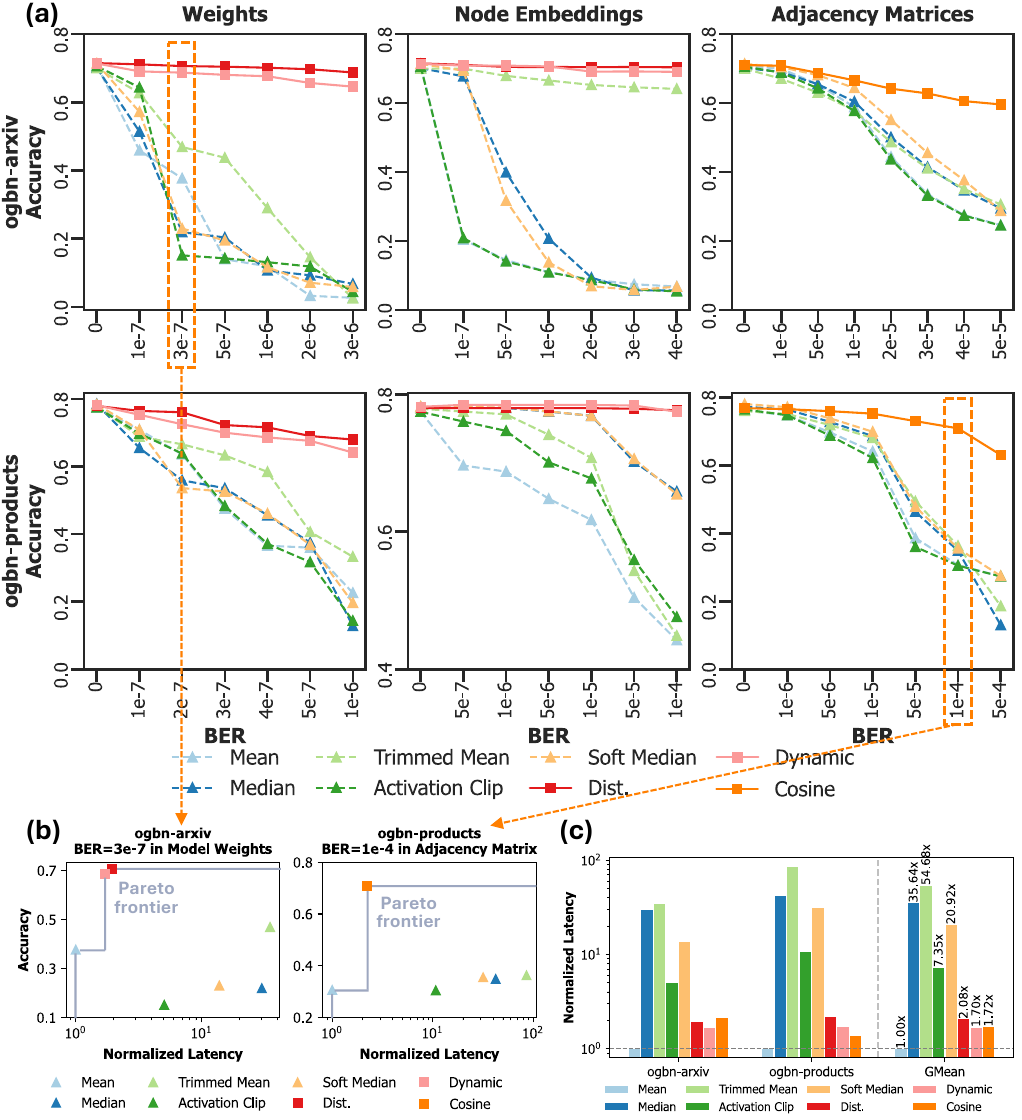}
    \caption{Comparison of GIN prediction accuracy and latency using robust aggregation functions versus existing methods on ogbn-arxiv and ogbn-products.
    (a) Prediction accuracy under varying BERs. Each row corresponds to a dataset, while each column represents bit-flip errors introduced in model weights, node embeddings, and adjacency matrices, respectively.
    (b) Pareto frontier of prediction accuracy versus normalized latency (relative to mean aggregation).
    (c) Profiled latency of different aggregation functions.}
    \label{fig:large}
\end{figure}

\textbf{Scalability to larger graphs.}
We further examine the robustness, computational complexity, and scalability of the proposed robust aggregation functions on larger and denser graph datasets, such as ogbn-arxiv and ogbn-products.
\begin{itemize}
    \item \textit{Robustness.} 
    As illustrated in Fig.~\ref{fig:large}(a), the proposed distribution-based, dynamic weight, and cosine aggregation functions consistently improve GNN robustness under bit-flip errors. This improvement holds across different graph scales. However, graphs with a larger number of nodes tend to exhibit lower tolerance to BERs, as the absolute volume of potentially corrupted data increases.
    \item \textit{Computational Complexity.} Table~\ref{tab:aggregation} summarizes the theoretical complexities of various aggregation strategies. Let $|E|$ be the number of edges and $l$ the dimension of node embeddings.
        \begin{itemize}
            \item Mean Aggregation: Standard mean aggregation has a time complexity of $O(l|E|)$~\cite{blakely2021time}.
            \item Median/Trimmed Mean Aggregation: These methods require element-wise sorting, resulting in a complexity of $O(l|E|\log|E|)$~\cite{liang2021understanding}.
            \item Soft Median Aggregation: Although it asymptotically scales as $O(l|E|)$ for small $l$~\cite{geisler2021robustness}, its use of softmax (and exponential) operations makes it more computationally intensive in practice.
            \item Our Proposed Robust Aggregation Functions: The distribution-based, dynamic weight, and cosine aggregations retain the same linear complexity as mean aggregation, $O(l|E|)$, ensuring efficient scalability.
        \end{itemize}
    \item \textit{Scalability.} 
    The runtime profiles in Fig.~\ref{fig:large}(c) show that median and trimmed mean incur significantly higher latency due to sorting operations, particularly in graphs with many high-degree nodes. While the activation clipping method is more efficient, it does not provide notable gains in robustness. Although soft median maintains linear complexity, its practical latency is higher due to the overhead of softmax operations. In contrast, our proposed aggregation functions deliver strong scalability and performance, with profiled latencies of 2.08$\times$, 1.70$\times$, and 1.72$\times$ that of mean aggregation for the distribution-based, dynamic weight, and cosine methods, respectively.
    \item \textit{Pareto frontiers.} 
    As shown in Fig.~\ref{fig:large}(b), our robust aggregation methods lie on favorable Pareto frontiers balancing accuracy and latency. One frontier is formed by the distribution-based and dynamic weight aggregations, and another by the cosine aggregation. Although some additional computation is inevitable for improved robustness, the proposed methods clearly outperform existing aggregations by offering a better trade-off between robustness and execution efficiency.
\end{itemize}

\begin{table}[t]
    \centering
    \caption{Time complexity analysis of different aggregation functions.}
    \begin{tabular}{cc|c}\toprule
    \multicolumn{2}{c|}{\textbf{Aggregation Function}} & \textbf{Time Complexity} \\ \midrule
    \multicolumn{1}{c|}{} & Mean & $O(l|E|)$ \\
    \multicolumn{1}{c|}{} & Activation Clip & $O(l|E|)$ \\
    \multicolumn{1}{c|}{} & Median & $O(l|E|\log{|E|})$ \\
    \multicolumn{1}{c|}{} & Trimmed Mean & $O(l|E|\log{|E|})$ \\
    \multicolumn{1}{c|}{\multirow{-5}{*}{\begin{tabular}[c]{@{}c@{}}Existing \\ Aggregation \\ Functions\end{tabular}}} & Soft Median & $O(l|E|)$ \\ \midrule
    \rowcolor[HTML]{DEECF7} 
    \multicolumn{1}{c|}{\cellcolor[HTML]{DEECF7}} & Distribution-based & $O(l|E|)$ \\
    \rowcolor[HTML]{DEECF7} 
    \multicolumn{1}{c|}{\cellcolor[HTML]{DEECF7}} & Dynamic Weight & $O(l|E|)$ \\
    \rowcolor[HTML]{DEECF7} 
    \multicolumn{1}{c|}{\multirow{-3}{*}{\cellcolor[HTML]{DEECF7}\begin{tabular}[c]{@{}c@{}}\textbf{\textit{Our Proposed}} \\ \textbf{\textit{Robust}} \\ \textbf{\textit{Aggregation}}\end{tabular}}} & Cosine & $O(l|E|)$ \\ \bottomrule
    \end{tabular}
    \label{tab:aggregation}
\end{table}


\vspace{10pt}
\section{Conclusion}
In this paper, we investigate the robustness of GNNs under bit-flip errors and observe that they generally remain resilient at BER below $10^{-7}$. However, larger graphs tend to tolerate lower BERs due to the increased absolute volume of data being affected. 
This insight suggests that for applications where reliability is not a primary concern, aggressive system design strategies can be adopted to improve energy and execution efficiency.

When GNNs are employed in safety-critical scenarios, we propose Ralts, robust aggregation techniques that leverage different graph similarity metrics and statistical information to mitigate the impact of extreme values, restore graph topology, and improve the overall resilience of GNNs.
On average, under a BER of $3\times10^{-5}$, distribution-based and dynamic weight aggregation functions improve the prediction accuracy by 44.1\%, 27.4\%, 32.6\%, 22.2\%, and 43.7\%, compared to mean, median, trimmed mean, soft median aggregation functions, and activation clipping; cosine aggregation improves prediction accuracy by around 10\% compared to mean/median/trimmed mean/soft median aggregation, and activation clipping.
In terms of execution efficiency, the latency of distribution-based, dynamic weight, and cosine aggregation functions incur 1.42$\times$, 1.08$\times$, and 1.37$\times$ the latency of the mean aggregation, respectively. 
Ralts scales well to denser and larger graphs, with the same asymptotic complexity as PyG built-in mean aggregation, and offers a favorable trade-off between robustness and computational efficiency.
Overall, Ralts consistently enhances GNN robustness across different GNN models, graph datasets, error patterns, and both dense and sparse architectures.


\bibliographystyle{ieeetr}
\bibliography{ref}

\end{document}